\documentclass[sigconf]{acmart}

\AtBeginDocument{%
  }

\copyrightyear{2026}
\acmYear{2026}
\setcopyright{cc}
\setcctype{by}
\acmConference[WWW Companion '26]{Companion Proceedings of the ACM Web Conference 2026}{April 13--17, 2026}{Dubai, United Arab Emirates}
\acmBooktitle{Companion Proceedings of the ACM Web Conference 2026 (WWW Companion '26), April 13--17, 2026, Dubai, United Arab Emirates}
\acmPrice{}
\acmDOI{10.1145/3774905.3795610}
\acmISBN{979-8-4007-2308-7/2026/04}

\begin{document}

\title{Robust Harmful Meme Detection under Missing Modalities via Shared Representation Learning}


\author{Felix Breiteneder}
\orcid{0009-0008-4351-5588}
\affiliation{%
  \institution{Johannes Kepler University}
  \city{Linz}
  \country{Austria}
}
\email{felix.breiteneder@aon.at}

\author{Mohammad Belal}
\orcid{0000-0002-6665-5873}

\affiliation{%
  \institution{Aalto University}
  \city{Espoo}
  \country{Finland}}
\email{mohammad.belal@aalto.fi}

\author{Muhammad Saad Saeed}
\orcid{0000-0002-0893-9499}

\affiliation{%
  \institution{University of Michigan-Flint}
  \city{Michigan}
  \country{USA}
}
\email{msaads@umich.edu}

\author{Shahed Masoudian}

\orcid{0009-0007-2747-0386}
\affiliation{%
 \institution{Johannes Kepler University}
 \city{Linz}
 \country{Austria}}
 \email{shahed.masoudian@jku.at}

\author{Usman Naseem}
\orcid{0000-0003-0191-7171}
\affiliation{%
  \institution{Macquarie University}
  \city{Sydney}
  \country{Australia}
}
\email{usman.naseem@mq.edu.au}

\author{Kulshrestha	Juhi}

\orcid{0000-0002-4375-4641}
\affiliation{%
  \institution{Aalto University}
  \city{Espoo}
  \country{Finland}}
\email{juhi.kulshrestha@aalto.fi}

\author{Markus Schedl}

\orcid{0000-0003-1706-3406}
\affiliation{
  \institution{Institute of Computational Perception, Johannes Kepler University Linz and Linz Institute of Technology}
  \city{Linz}
  \country{Austria}
}
\email{markus.schedl@jku.at}

\author{Shah Nawaz}
\orcid{0000-0002-7715-4409}

\affiliation{%
  \institution{Johannes Kepler University}
  \city{Linz}
  \country{Austria}
}
\email{shah.nawaz@jku.at}
\renewcommand{\shortauthors}{Breiteneder et al.}

\begin{abstract}
Internet memes are powerful tools for communication, capable of spreading political, psychological, and sociocultural ideas. However, they can be harmful and can be used to disseminate hate toward targeted individuals or groups. Although previous studies have focused on designing new detection methods, these often rely on modal-complete data, such as text and images.  In real-world settings, however, modalities like text may be missing due to issues like poor OCR quality, making existing methods sensitive to missing information and leading to performance deterioration. To address this gap, in this paper, we present the first-of-its-kind work to comprehensively investigate the behavior of harmful meme detection methods in the presence of modal-incomplete data.  Specifically, we propose a new baseline method that learns a shared representation for multiple modalities by projecting them independently. These shared representations can then be leveraged when data is modal-incomplete. 
Experimental results on two benchmark datasets demonstrate that our method outperforms existing approaches when text is missing. Moreover, these results suggest that our method allows for better integration of visual features, reducing dependence on text and improving robustness in scenarios where textual information is missing. 
Our work represents a significant step forward in enabling the real-world application of harmful meme detection, particularly in situations where a modality is absent.
\end{abstract}

\ccsdesc[500]{Computing methodologies~Neural networks}

\keywords{Multimodal Learning, Vision and Language, Harmful Memes, Missing Modalities}



\maketitle

\section{Introduction}

In the digital era, memes have become a prevalent form of communication, blending images and text to convey humor, satire, or sociocultural ideas. Their rapid spread across social media platforms highlights their influence on public discourse. While many memes are benign or satirical, a growing subset is being used to spread hate, misinformation, or harassment, targeting individuals, social groups, or institutions~\cite{alam2021survey}. For instance, harmful memes may embed extremist ideologies, promote self-harm, or perpetuate stereotypes, leveraging the interplay of visual and textual cues to evade traditional content moderation methods~\cite{naseem2023multimodal,chen2024unveiling}. The proliferation of such content shows the urgency of developing robust detection mechanisms, particularly as platforms grapple with balancing free expression and user safety~\cite{alam2021survey}.  Detecting harmful memes presents challenges due to their multimodal nature, combining visual and textual components that often convey implicit or context-based meanings~\cite{pramanick2021detecting}.

\begin{table}[!t]
\centering
  \caption{Comparison of the performance of OCR models on test set of combined Harm-C and Harm-P datasets based on Word Error Rate (WER) and BiLingual Evaluation Understudy) BLEU scores. The performance of these models on real world memes data highlights the unreliability of text availability in practice.}
\centering
  \small 
\resizebox{.85\columnwidth}{!}{
  \begin{tabular}{lcc}
    \toprule
    \textbf{Model} & \textbf{Word Error Rate (WER)} & \textbf{BLEU} \\
    \midrule
    EasyOCR    & 0.5522  & 0.4617 \\
    PaddleOCR  & 0.5197  & 0.4711 \\
    Tesseract  & 0.8129  & 0.1969 \\
    \bottomrule
  \end{tabular}
  }

    \label{tab:ocr_evaluation}
    \vspace{-5mm}
\end{table}

Traditional methods have utilized machine learning techniques to analyze these two components, separately or in combination, to identify harmful content \cite{pramanick2021, zhou2021multimodal, Velioglu2020DetectingHS,NEURIPS2020_1b84c4ce}. Existing harmful meme detection methods extensively rely on multimodal frameworks that analyze both visual and textual elements, such as VisualBERT~\cite{li2019visualbert}, CLIP-based models, or cross-modal attention fusion techniques. For example, the MOMENTA~\cite{pramanick2021} framework employs a multimodal network that examines both global and local perspectives of a meme to assess its potential harm. This approach underscores the importance of considering the interplay between text and image to understand the message conveyed by memes. These approaches often achieve strong performance in controlled settings, as evidenced by benchmarks, like the Hateful Memes Challenge~\cite{kiela2020hateful}, which emphasizes the benefit of joint modality analysis. Recent research has begun to address these challenges by incorporating large language models into the detection process \cite{lin-etal-2023-beneath, huang-etal-2024-towards, capAlign}.

However, a critical limitation arises in real-world scenarios where one modality (e.g., text) may be missing, corrupted, or intentionally obfuscated to bypass detection. For example, memes might use heavily modified images with minimal text or rely on cultural subtexts invisible to automated systems \cite{ijcai2022p781}. Current methods, which assume modal completeness, exhibit sensitivity to such gaps, leading to deteriorated performance and reliability. 
See Table~\ref{tab:ocr_evaluation} for a preliminary analysis showcasing that OCR systems often fail to extract the embedded text in memes, validating our hypothesis that text availability is often unreliable in real-world scenarios.

In this work, we propose a new baseline method for harmful meme detection that learns a shared representation for multiple modalities by independently projecting each modality. This method focuses on real-world applicability, where memes are often ambiguous, missing information or a modality, and aims to provide a more adaptable framework for detecting harmful content in diverse digital environments. Our contributions are as follows:
\begin{itemize}
    \item To the best of our knowledge, this is the first work to comprehensively study the problem of harmful meme detection under the realistic condition of missing text modality.
    \item We propose a baseline method that learns a shared representation for multiple modalities, enabling robust performance even with modal-incomplete data.
    \item We present extensive experimental results and analysis, demonstrating the effectiveness of our proposed method in handling missing text, highlighting its improved robustness and real-world applicability.
\end{itemize}

\section{Related Work}

\subsection{Harmful Meme Detection}
Prior work on harmful meme detection are focused on detecting misinformation, hate speech, and offensive content in multimodal settings. Several shared tasks, including troll meme classification \cite{Suryawanshi2021FindingsOT} and hateful meme detection \cite{kiela2020hateful, zhou2021multimodal} have played a role in advancing the task. 
While early methods explored logistic regression and BERT-based classifiers \cite{Devlin2019BERTPO}, recent work has focused on multimodal networks that jointly process text and images. Models such as VisualBERT \cite{li2019visualbert}, UNITER \cite{Chen2019UNITERUI}, ViLBERT \cite{Lu2019ViLBERTPT}, and LXMERT \cite{Tan2019LXMERTLC} have shown improved performance by leveraging joint vision-language representations.
Further studies have used multimodal deep ensembles \cite{Sandulescu2020DetectingHM} to improve model performance. However, these approaches focus more on hate speech detection and sentiment analysis rather than addressing harmful meme detection as a whole.

Modality alignment remains a significant obstacle in harmful meme detection, as textual and visual components often contribute complementary information. Different fusion techniques have been developed to address this issue. Some methods combine encoded representations from different modalities into a shared feature space \cite{Poria2016ConvolutionalMB, Wang2016SelectadditiveLI}, while others process each modality independently before merging the representations \cite{zhou2021multimodal}. MOMENTA \cite{pramanick2021} introduced a cross-modal attention mechanism to improve alignment between meme images and background text, demonstrating improved performance over traditional fusion techniques.

Other studies have investigated modality transfer as an alternative to direct fusion. Pro-Cap \cite{Cao2023ProCapLA} employs pretrained vision-language models in a zero-shot visual question answering framework to generate image captions for harmful content detection. Similarly, \cite{Promptbasedapproach} proposed converting multimodal meme content into textual captions, performing well on benchmark datasets. However, these methods rely on high-quality captioning and may not generalize well to complex multimodal interactions.   

CapAlign \cite{capAlign} extends this line of research by generating informative captions rather than generic textual descriptions. By prompting a large language model to ask targeted questions to a pretrained vision-language model, CapAlign extracts contextually relevant details before aligning the generated caption with the textual content of a meme. This method improves cross-modal alignment and has demonstrated state-of-the-art performance on harmful meme detection benchmarks.

\begin{figure*}[t]
    \centering
    \includegraphics[width=0.99\textwidth]{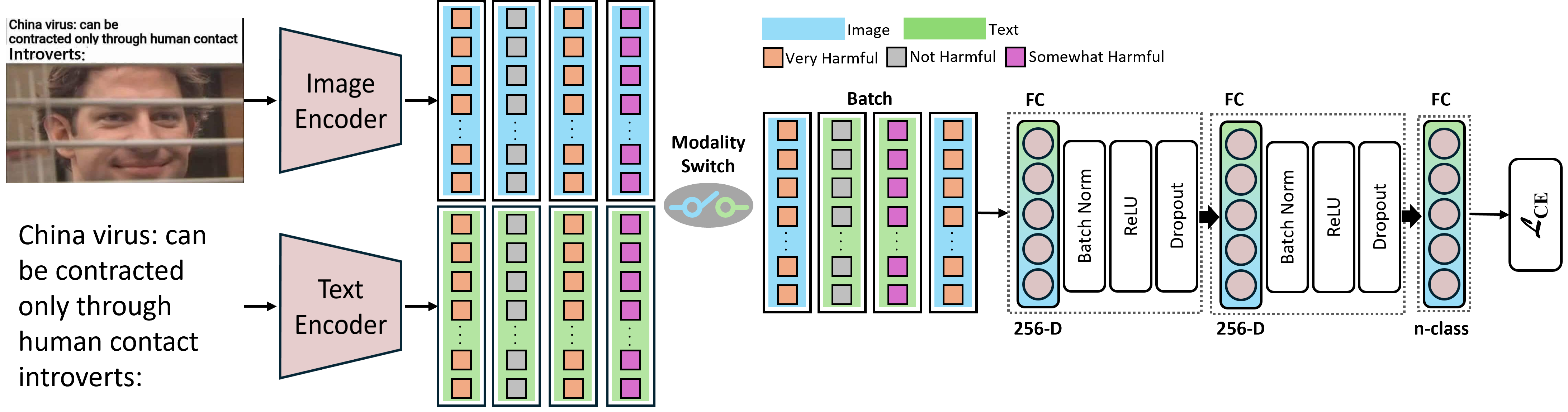}  
    \caption{Overview of our model. Each modality is independently selected through a switch and passed through a network of fully-connected layers to learn shared representations.}
    \label{fig:architecture}
\end{figure*}

\subsection{Missing Modalities} 
Multimodal learning integrates information from different modalities, such as natural language, visual signals, and audio, to improve performance across various tasks. A key challenge in this field is fusion, where different modalities are combined to enhance model understanding. Simple fusion methods, such as feature concatenation, have been widely used \cite{Poria2016ConvolutionalMB, Wang2016SelectadditiveLI}, while more sophisticated approaches like tensor fusion \cite{Zadeh2017TensorFN} and low-rank fusion \cite{Liu2018EfficientLM} address scalability issues. However, these methods assume the availability of all modalities, which is rarely the case in real-world scenarios~\cite{liaqat2025multimodal}.

To improve robustness of multimodal learning methods, \cite{multimotransf} investigated the sensitivity of transformers to missing modalities and proposed an adaptive search-based optimization approach to dynamically select the most effective fusion strategy. This method mitigates performance degradation by adapting fusion mechanisms based on training conditions.
Other approaches focus on ensuring robustness when modalities are incomplete \cite{Ma_2022_CVPR, Ma2021SMILML, tagassisted, zhao-etal-2021-missing,liaqat2025chameleon,saad2024modality}. SMIL \cite{Ma2021SMILML} estimates latent features via Bayesian Meta-Learning, while \cite{tagassisted} introduced a tag encoding module to adapt transformers dynamically. MMIN \cite{zhao-etal-2021-missing} predicts missing modality representations from available data, and Ma et al.~\cite{Ma_2022_CVPR} explored multi-task optimization to improve fusion strategies in incomplete scenarios.
Lee et al.~\cite{Lee2023MultimodalPW} introduced modality-aware prompts as an alternative to full model retraining. This approach leverages prompt learning to efficiently adapt transformers, where input-level prompts improve performance, while attention-level prompts enhance stability across different datasets.

\section{Methodology}
\subsection{Problem Definition}

Harmful meme detection is a multimodal task that requires jointly analyzing textual and visual information to determine whether a meme contains harmful content. Formally, given a dataset $\mathcal{D} = \{(x_t, x_v, y)\}_{i=1}^{N}$, where $x_t$ and $x_v$ represent the text and image modalities respectively, and $y \in \mathcal{Y}$ is the target label, the goal is to learn a function $f: (x_t, x_v) \rightarrow y$ that accurately predicts the category of a meme. 
Existing multimodal methods typically adopt a two-branch network, where separate branches process text and image data before combining them to learn fused representations for classification task. While effective and modular under full-modality conditions, these models rely heavily on the availability of modalities and experience dramatic performance deterioration when a modality is missing. Since real-world scenarios often involve incomplete or noisy data, such as memes with unreadable or missing text, models must be designed to effectively handle it.
In this work, we address this challenge by introducing a multimodal learning method that processes image and text independently to learn a shared representation. 
As a result, the model remains robust when text is unavailable, shared representation of cross-modal interactions to compensate for missing modality data. 
Our method improves classification performance under missing modality scenario while maintaining competitive performance when both modalities are present.

Our proposed model (Figure~\ref{fig:architecture}) employs a an approach where both textual and visual modalities embeddings are mapped independently into a shared latent space. 
Unlike traditional multimodal transformers that rely on independent branches for each modality, our approach ensures that textual and visual features are projected into a shared representation space, improving robustness under missing modality conditions.

\subsection{Modality Encoder}

We use pre-extracted CLIP embeddings for both modalities, where $x_t \in \mathbb{R}^{d}$ represents the textual embedding and $x_v \in \mathbb{R}^{d}$ represents the visual embedding. These embeddings serve as input to a shared processing module.

\subsection{Shared Representation Network}
It comprises of two stacked blocks, each consisting of a fully connected (FC) layer followed by Batch Normalization (BN), ReLU activation, and dropout. This design enables the model to learn shared representations by independently processing text and image embeddings through the same set of parameters.
When both modalities are available at test time, their corresponding logits are summed before applying softmax for the final classification. If only one modality is present at test time, the model directly uses the available modality’s logits without requiring modality-specific adjustments.

\subsection{Loss Function}
We optimize the model using a cross-entropy loss over the predicted class distribution. 
By processing both modalities within a shared representation space, our model effectively learns a cross-modal embedding that enables it to retain strong classification performance even when textual data is missing.

\begin{table*}[!t]
  \caption{F1-scores for binary classification across Harm-C and Harm-P. Highest scores are bold and second highest are underlined.}
  \centering
  \begin{tabular}{l|ccccccc|ccccccc}
    \toprule
    & \multicolumn{7}{c|}{\textbf{Harm-C (Binary)}} & \multicolumn{7}{c}{\textbf{Harm-P (Binary)}} \\
    \textbf{Model} & \textbf{100} & \textbf{90} & \textbf{70} & \textbf{50} & \textbf{30} & \textbf{10} & \textbf{0} & \textbf{100} & \textbf{90} & \textbf{70} & \textbf{50} & \textbf{30} & \textbf{10} & \textbf{0} \\
    \midrule
    VisualBERT~\cite{li2019visualbert} & 77.6 & \underline{76.0} &\underline{ 71.2} & 67.5 & \underline{61.2} & 55.0 & 53.9 & 61.0 & 60.0 & \underline{58.0} & 54.5 & 48.2 & 39.3 & 11.8 \\
    MMBT~\cite{kiela2019supervised} & 78.0 & \underline{76.0} & 71.0 & \underline{68.0} & 61.0 & \underline{57.0} & \underline{55.0} & \underline{64.9} & \underline{61.9} & 55.9 & 50.1 & 38.4 & 21.7 & 11.8 \\
    MOMENTA~\cite{pramanick2021} & \textbf{80.8} & \textbf{77.5} & 70.4 & 63.5 & 55.7 & 43.0 & 38.1 & 59.8 & 58.4 & 56.9 & \underline{55.5} & \underline{52.3} & \underline{47.9} & \underline{45.5} \\
    Ours & \underline{78.7} & 75.6 & \textbf{71.7} & \textbf{68.1} & \textbf{65.3} & \textbf{61.0} & \textbf{59.1} & \textbf{67.5} & \textbf{67.4} & \textbf{67.3} & \textbf{67.0} & \textbf{66.7} & \textbf{66.5} & \textbf{66.3} \\
    \bottomrule
  \end{tabular}

  \label{tab:binary_results}
\end{table*}

\begin{table*}[htbp]
  \caption{F1-scores for multi-class classification across Harm-C and Harm-P. Highest scores are bold and second highest are underlined.}
  \centering
  \begin{tabular}{l|ccccccc|ccccccc}
    \toprule
    & \multicolumn{7}{c|}{\textbf{Harm-C (Multi-class)}} & \multicolumn{7}{c}{\textbf{Harm-P (Multiclass)}} \\
    \textbf{Model} & \textbf{100} & \textbf{90} & \textbf{70} & \textbf{50} & \textbf{30} & \textbf{10} & \textbf{0} & \textbf{100} & \textbf{90} & \textbf{70} & \textbf{50} & \textbf{30} & \textbf{10} & \textbf{0} \\
    \midrule
    VisualBERT~\cite{li2019visualbert} & 49.1 & 47.4 & 45.6 & 42.3 & \underline{40.6} & \underline{38.6} & \underline{38.1} & 35.6 & 35.1 & 34.8 & 34.7 & 32.5 & \underline{28.9} & \underline{28.4} \\
    MMBT~\cite{kiela2019supervised} & \textbf{54.4} & \underline{53.4} & \underline{50.7} & \underline{44.7} & 37.4 & 31.7 & 29.0 & \textbf{47.1} & \underline{42.7} & \underline{40.3} & \underline{38.9} & \underline{31.7} & 25.8 & 22.4 \\
    MOMENTA~\cite{pramanick2021} & 52.1 & 49.3 & 43.6 & 37.6 & 30.6 & 21.9 & 16.6 & 39.0 & 37.9 & 35.1 & 33.0 & 31.1 & 26.6 & 25.9 \\
    Ours & \underline{54.2} & \textbf{53.4} & \textbf{53.4} & \textbf{52.9} & \textbf{52.0} & \textbf{51.3} & \textbf{51.1} & \underline{46.1} & \textbf{45.6} & \textbf{45.6} & \textbf{43.8} & \textbf{43.1} & \textbf{42.4} & \textbf{42.4} \\
    \bottomrule
  \end{tabular}

  \label{tab:multiclass_results}
\end{table*}

\section{Experimental Settings}
\subsection{Dataset}

We use the HarMeme dataset for our experiments, which consists of two subsets: Harm-C (memes related to COVID-19) and Harm-P (political memes). The dataset annotates each meme into one of three categories: very harmful, somewhat harmful, or not harmful. These categories define both binary and multi-class tasks.
To the best of our knowledge, this is the first study reporting performance metrics on the corrected Harm-P dataset. The original dataset, introduced by \cite{pramanick2021}, contained duplicate samples across the training and test/validation splits, leading to data leakage. The authors later recognized this issue and released an updated version, which we use in our experiments.

\subsection{Baselines}
\label{sec:baselines}
We compare our model against three established multimodal methods: MMBT~\cite{kiela2019supervised}, VisualBERT~\cite{li2019visualbert}, and MOMENTA~\cite{pramanick2021}.

\begin{itemize}
    \item \textbf{MMBT} \cite{kiela2019supervised} processes text using a pretrained BERT model and images using a ResNet backbone. The extracted features are concatenated and passed through a transformer encoder, enabling late fusion of multimodal data.

    \item \textbf{VisualBERT} \cite{li2019visualbert} integrates textual and visual representations by injecting image region embeddings into a BERT-based transformer. The model is pretrained on vision-language tasks, learning aligned  representations through early fusion.

    \item \textbf{MOMENTA} \cite{pramanick2021} is designed for harmful meme detection, encoding both global and local multimodal features. It extends a transformer-based multimodal classifier with an auxiliary contrastive loss to enhance the distinction between harmful and non-harmful memes.
    
\end{itemize}

\subsection{Implementation Details}
We train our model on an NVIDIA A$100$ GPU for $5$ epochs using a batch size of $8$. Training uses AdamW optimizer with an initial learning rate of $1e^{-4}$ and gradient clipping at $2.0$. The learning rate schedule includes a linear warm-up over the first $20$\% of the training steps, followed by a linear decay to $10$\% of the initial rate. All experiments are conducted with $5$-fold cross-validation and repeated with 3 different random seeds. We adopt the established train, validation, and test splits \cite{pramanick2021}.

\subsection{Evaluation Tasks}
We evaluate our method on two classification tasks: binary classification and multi-class classification, as well as in conditions where text input is missing for a subset of samples.
In the binary classification task, the very harmful and somewhat harmful categories are merged into a single harmful class. The multi-class classification task retains the original three labels: very harmful, somewhat harmful, and not harmful.

To assess model performance when text input is missing for a portion of the test set, we conduct missing modality experiments. We evaluate models at different levels of text availability: $100$\%, $90$\%, $70$\%, $50$\%, $30$\%, $10$\%, and $0$\%. At $10$\% text availability, for example, only 10\% of test samples contain text and image data, while the remaining 90\% are composed of the image modality. 
The selection of samples with text is controlled to maintain the same label distribution as the full test set, ensuring a consistent evaluation.
To account for class imbalance, we use the F1-score as the primary evaluation metric. Models are selected based on validation performance, and all results are reported on the test set of the corrected HarMeme dataset.

\section{Results and Analysis}
\subsection{Binary Classification}
Table~\ref{tab:binary_results} shows that our model outperforms all baselines on Harm-P across all text availability levels. On Harm-C, MOMENTA achieves the highest score when text is fully available, but its performance degrades sharply as text availability decreases. Our model surpasses all baselines when text availability is $70$\% or lower, demonstrating robustness to missing text.

\subsection{Multi-class Classification}

In Table~\ref{tab:multiclass_results}, our model achieves the best performance on Harm-C and Harm-P when the availability of text is $90$\% or lower. Although MMBT performs best with full text availability, its performance decreases significantly as text is removed. VisualBERT consistently ranks lower than other baselines, especially struggling under missing text conditions. In contrast, our model remains more stable across different text availability levels, highlighting the ability of weight sharing to improve modality robustness.

\subsection{Impact of Missing Text}
We observed that performance drops dramatically for all models when text availability is $30$\% or lower, with F1-scores deteriorating up to $42$\% in some cases. MOMENTA and MMBT show the steepest declines, indicating a strong reliance on text. In contrast, our model retains a higher degree of stability, maintaining a competitive F1-score even at $0$\% text availability. This suggests that our proposed model allows for better integration of visual features, reducing dependence on text, and improving robustness in scenarios where textual information is unreliable.

\begin{table}[t]
  \centering
  \caption{F1-scores for multi-class classification across Harm-C. Highest scores are bold and second highest are underlined.}
  \begin{tabular}{l|c|c}
    \toprule
    & \multicolumn{2}{c}{\textbf{Harm-C (Multi-class)}} \\
    \textbf{Model}   & Original & Noisy  \\
    \midrule
    MOMENTA~\cite{pramanick2021}  & \underline{52.1} & \underline{40.8} \\
    \bottomrule 
    Ours    & \textbf{54.2} & \textbf{50.0} \\
    \bottomrule
  \end{tabular}
  \label{tab:noisy_text}
\end{table}

\subsection{Impact of Noisy Text}
We also compared the performance of our model on MOMENTA with noisy text extracted from PaddleOCR using Harm-C multi-class configuration. In Table~\ref{tab:noisy_text}, MOMENTA shows dramatic deterioration, indicating a strong reliance on high quality text descriptions. In contrast, our model retains a higher degree of stability.

\subsection{Ablation Study}

Existing multimodal learning methods for detecting harmful memes learn shared representations by mapping the fused embeddings of multiple modalities. These fused representations (FR) provides overlapping and complementary information suitable to perform the task when modalities are available. In contrast, our baseline method that learns a shared representation (SR) for multiple modalities by mapping them individually in the embedding space. These shared representations can then be leveraged when data is modal-incomplete. 
Table~\ref{tab:ablation_network} compared fused and shard representations configurations with SOTA methods. 
These results demonstrate that shared representations perform slightly lower compared to fused representations when modalities are completely available. However, shared representations dramatically outperforms fused representation when the textual modality is missing, demonstrating the effectiveness of learning to map individual modality.

\begin{table}[!t]
  \caption{F1-scores for binary classification across Harm-C. Highest scores are bold and second highest are underlined.}
  \centering
  \begin{tabular}{l|cccc}
    \toprule
    & \multicolumn{4}{c}{\textbf{Harm-C (Binary)}} \\
    \textbf{Model} & \textbf{100} & \textbf{70} & \textbf{50} & \textbf{10}  \\
    \midrule
    VisualBERT~\cite{li2019visualbert}     & 77.6  &\underline{ 71.2} & 67.5 & 55.0 \\
    MMBT~\cite{kiela2019supervised}           & 78.0  & 71.0 & \underline{68.0} & \underline{57.0}  \\
    MOMENTA~\cite{pramanick2021} & \textbf{80.8}  & 70.4 & 63.5 & 43.0  \\
    \bottomrule 
    Ours\_{FR}   & \textbf{80.8}  & 66.8 & 56.3 & 17.3  \\
    Ours\_{SR}   & \underline{78.7}  & \textbf{71.7} & \textbf{68.1} & \textbf{61.0}  \\
    \bottomrule
  \end{tabular}

  \label{tab:ablation_network}
\end{table}

\section{Conclusion}
We empirically find that harmful meme detection methods are sensitive to modal-incomplete data. 
Based on these findings, we introduce a novel baseline method that learns a shared representation across multiple modalities by independently mapping each modality in the embedding space. These shared representations prove effective in handling scenarios with complete and missing data.
We believe our work represents a significant step forward in enabling real-world applications of harmful meme detection, specifically in scenarios where a certain modality is missing.

\section{Acknowledgments}
This research was funded in whole or in part by the Austrian Science Fund (FWF): Cluster of Excellence \href{https://www.bilateral-ai.net/home}{\textcolor{blue}{\textit{Bilateral Artificial Intelligence}}} (\url{https://doi.org/10.55776/COE12}), the doc.funds.connect project \href{https://dfc.hcai.at/}{\textcolor{blue}{\textit{Human-Centered Artificial Intelligence}}} (\url{https://doi.org/10.55776/DFH23}), and the PI project \href{https://doi.org/10.55776/P36413}{\textcolor{blue}{\textit{Intent-aware Music Recommender Systems}}} (\url{https://doi.org/10.55776/P36413}).
For open access purposes, the authors have applied a CC BY public copyright license to any author-accepted manuscript version arising from this submission.










\balance
\bibliographystyle{ACM-Reference-Format}
\bibliography{sample-base}


\end{document}